\documentclass{article}

\usepackage{microtype}
\usepackage{graphicx}
\usepackage{subcaption}
\usepackage{booktabs} 

\usepackage{hyperref}

\usepackage[preprint]{icml2026}

\usepackage{amsmath}
\usepackage{amssymb}
\usepackage{mathtools}
\usepackage{amsthm}
\usepackage{natbib}

\usepackage[capitalize,noabbrev]{cleveref}

\theoremstyle{plain}

\theoremstyle{definition}

\theoremstyle{remark}

\usepackage[textsize=tiny]{todonotes}

\icmltitlerunning{Directional Concentration Uncertainty}

\begin{document}

\twocolumn[
\icmltitle{Directional Concentration Uncertainty: A representational approach to uncertainty quantification for generative models}




\begin{icmlauthorlist}
\icmlauthor{Souradeep Chattopadhyay}{ui}
\icmlauthor{Brendan Kennedy}{pnnl}
\icmlauthor{Sai Munikoti}{pnnl}
\icmlauthor{Soumik Sarkar}{ui}
\icmlauthor{Karl Pazdernik}{pnnl,ncsu}
\end{icmlauthorlist}

\icmlaffiliation{ui}{Department of Mechanical Engineering, Iowa State University, Ames, IA, USA}
\icmlaffiliation{pnnl}{Pacific Northwest National Laboratory, Richland, WA, USA}
\icmlaffiliation{ncsu}{Department of Statistics, North Carolina State University, Raleigh, NC, USA}

\icmlcorrespondingauthor{Souradeep Chattopadhyay}{soura@iastate.edu}
\icmlcorrespondingauthor{Brendan Kennedy}{brendan.kennedy@pnnl.gov}

  \icmlkeywords{Machine Learning, ICML}

  \vskip 0.3in
]



\printAffiliationsAndNotice{}  
\begin{abstract}
In the critical task of making generative models trustworthy and robust, methods for Uncertainty Quantification (UQ) have begun to show encouraging potential. However, many of these methods rely on rigid heuristics that fail to generalize across tasks and modalities. Here, we propose a novel framework for UQ that is highly flexible and approaches or surpasses the performance of prior heuristic methods.
We introduce \emph{Directional Concentration Uncertainty} (DCU), a novel statistical procedure for quantifying the concentration of embeddings based on the von Mises-Fisher (vMF) distribution. Our method captures uncertainty by measuring the geometric dispersion of multiple generated outputs from a language model using continuous embeddings of the generated outputs without any task specific heuristics. 
In our experiments, we show that DCU matches or exceeds calibration levels of prior works like semantic entropy \citep{kuhnetal23} and also generalizes well to more complex tasks in multi-modal domains. We present a framework for the wider potential of DCU and its implications for integration into UQ for multi-modal and agentic frameworks.
\end{abstract}
\section{Introduction}

Despite their obvious utility, generative models such as Large Language Models (LLMs) cannot be deployed in sensitive domains or used directly in decision making if they are unable to provide reliable, transparent outputs. There are myriad frameworks proposed by the community to address this limitation, including interpretability frameworks \citep{palikhe2025towards,yu2024mechanistic} and validation via external knowledge sources \citep{agrawal2024can,sui2025can}. One family of techniques that is now increasingly studied, yet still underdeveloped, is uncertainty quantification (UQ). 

UQ is a layer between model and user that can increase transparency and detect unreliability. By exposing aspects of the models' underlying confidence (or lack thereof), UQ can give users sufficient understanding of the underlying model processes such that they can make informed downstream decisions given model outputs.

UQ has a rich history in both conventional modeling and the more recent deep learning wave, with techniques for the latter including Bayesian neural networks \citep{neal1992bayesian}, neural network uncertainty via dropout \citep{gal2016dropout}, and conformal prediction \citep{vovk2005algorithmic,shafer2008tutorial}. UQ for LLMs is a newer research area that has required a different set of approaches, due in large part to the computational complexity associated with the sheer size of models and the nature of discrete, sequential generations \citep{liu2025uncertainty}. 
Among these more recent directions, one of the most important is the design of black-box algorithms that rely only on model outputs, allowing application to closed-source models. 
In this area, most approaches have a sampling-based strategy, drawing multiple high-temperature samples and characterizing variance in the output samples \citep{kuhnetal23,aichberger2024semantically,quachconformal,stengel2024lacie,ulmer2024non}.

Of note among these sampling-based methods is semantic entropy (SE), which achieves high calibration scores for model correctness on question-answering tasks \cite{kuhnetal23} and as a black-box hallucination detection method \cite{farquharetal24}. 
After sampling from a model, SE proceeds with two steps: (1) clustering responses based on semantic equivalence, operationalized using entailment, and (2) computing the entropy over cluster assignment probabilities (see Section~\ref{sec:method}).

We observe that a number of factors limit the true effectiveness and generality of SE. The success of SE and competing sampling-based methods has been demonstrated through evaluation on conventional, structured question-answering datasets, such as TriviaQA \citep{triviaqa}. When applied to more challenging, open-ended tasks, we find that the performance of SE suffers significantly (see Section~\ref{sec:results:visual}).  
We attribute this to a structural limitation of SE: the clustering of model responses requires a narrowly defined notion of semantic equivalence. Specifically, the notion of textual entailment --- determining that the meaning of one text can be inferred from another text --- works effectively for short responses in the question-answering domain, but cannot be applied to other types of text, particularly those that are longer and more complex.
This is due to the ad hoc assumptions which are necessary for comparing pieces of text, which can vary significantly by task type (e.g., question answering versus summarization), task complexity, and modality. Overall, approaches relying on text clustering will struggle to generalize across all language domains.

We propose \textbf{Directional Consistency Uncertainty} (DCU), a novel framework that expands on the success of semantic entropy and other sampling-based approaches in order to achieve a more general and flexible UQ for generative models. DCU is a geometric framework that relies on embedded representations of samples from a model. 
DCU models the set of sampled outputs as a distribution in a representation space and quantifies how directionally concentrated that distribution is. This directional alignment reflects consistent generation (low uncertainty), whereas substantial deviations reflect variability in the model’s responses (high uncertainty).

In our experiments, we perform comparisons between SE and DCU on established question-answering benchmarks, showing that our flexible, embedding-based approach can perform acceptably at the same tasks on which SE with semantic equivalence excels. Then, we perform experiments on a more complex domain, visual question-answering, demonstrating the performance limitations of SE on this task and the improvement in performance of DCU. We conclude by presenting a vision for the wider application of DCU as a framework for capturing variability in model responses across modalities and tasks, as well as in agentic systems.

\section{Related Work}

UQ for LLMs can be broadly approached through two approaches, (1) model based and (2) sampling based. \cite{xialetal22} performed a comprehensive survey of different UQ methods LLMs for both approaches on a variety of tasks like text classification, text generation, question answering etc. They evaluated several popular UQ methods like Bayesian neural networks, ensembling etc. using popular datasets like IMBD AG News etc. Through their investigations they made suggestions about use of different UQ methods for different tasks also also discussed the possible limitations of the different UQ methods they investigated.

\cite{linetal24} proposed a method that can assess UQ of LLMs without access to internal gradients or logits of the models. They explored their approach using popular models like GPT 3.5 turbo, LLaMA,  using datasets like TriviaQA. For evaluation they used simple metrics like calibration curves and expected calibration error. Their method showed significant improvement in terms of calibration (difference between actual probability of correctness and predicted probability of correctness) but also suffered from drawbacks like computational costs and generalization. \cite{lingetal24} proposed a method of uncertainty quantification for in-context learning tasks that uses the concept of entropy decomposition to quantify both aleatoric and epistemic uncertainty in an unsupervised way. \cite{yeetal24} proposed a new approach of benchmarking LLMs that address the issue of uncertainty unlike popular open leaderboards like HuggingFace.

Several other methods have been proposed recently that uses the concepts like semantic uncertainty to propose UQ methods for LLMs. \citet{kuhnetal23} introduced a metric called semantic entropy that incorporate semantic equivalence a.k.a the idea that different sentences that carry the same meaning. Their proposed metric showed incredible performance for different natural language generation tasks such as question answering. Following this other methods like \citet{qiuetal24} were proposed that incorporate the concept of semantic uncertainty. \citet{qiuetal24} introduced the idea of semantic density, a metric to quantify the confidence of LLM responses in a semantic space by probability distributions of the LLM outputs from a semantic perspective. Their proposed method evaluates trustworthiness of the LLM outputs, considers the concept of semantic differences among outputs and is also free of additional training or fine tuning. Also their approach performed better compared to prior approaches on different question answering tasks using state of thr art models like LLaMA 3.

Some other recent development in this domain involves use of perturbation methods to measure UQ for LLMs. \citet{gaoetal24} introduced a novel method call sampling with perturbation for uncertainty quantification (SPUQ) capable of handling both aleatoric and epistemic uncertainties by generating perturbations in the LLM inputs followed by aggregation of sampling outputs from each perturbation. They experiment using different state of the art models like GPT 3, GPT 4, PaLM2 and different datasets like XSUM, StrategyQA for text generation tasks. Their proposed approach exhibited reduction of expected calibration error in the range of 30-70\%. Apart from perturbation methods other methods have been proposed to quantify uncertainty of LLMs.  
\section{Method} \label{sec:method}
\begin{figure*}[t]
    \centering
    \includegraphics[width=\textwidth]{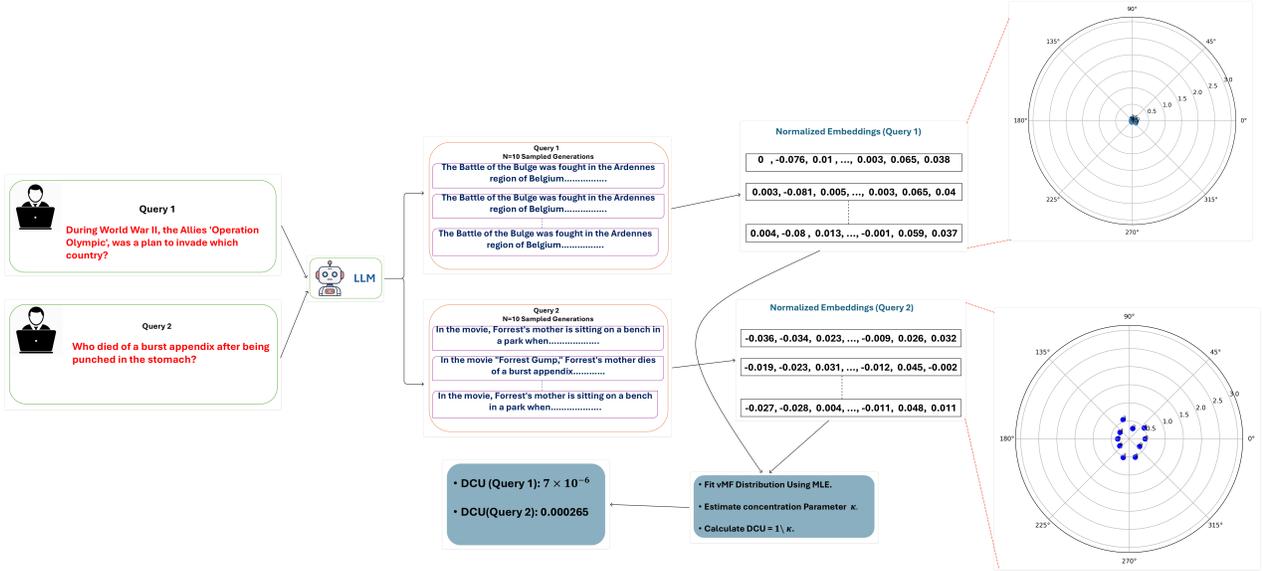}
    \caption{A schematic representation of the directional consistency measure using two LLM queries from TriviaQA evaluated with LLaMA-2. For each query, 10 generations were sampled and encoded using the e5-large-v2 model to obtain embeddings. The radial plots illustrate the angular dispersion of the embeddings relative to the estimated mean direction.}
    \label{fig:dcu_ex}
\end{figure*}

Here, we present our geometric method, Directional Concentration Uncertainty, clarify its differences with the clustering-based semantic entropy method, and describe how we derive an uncertainty measurement using the von Mises-Fisher (vMF) distribution.

\subsection{Problem Formulation}
For our problem we considered a setup where a language model was queried with an input $x$ resulting in a textual output $y_x$. Our objective is to quantify the \textit{uncertainty} of $y$, defined as the variability in the content of multiple outputs generated for $x$.

Similar to SE, our approach estimates uncertainty by analyzing the variability across multiple sampled outputs for the same input. However, unlike SE, we do not rely on semantic clustering or bidirectional entailment models, and instead operate directly on continuous embedding representations of the model's output. More details about the difference between our approach and SE are provided in the next section.

\subsubsection{Differentiation with Semantic Entropy}

SE estimates uncertainty by grouping multiple sampled outputs into semantically equivalent clusters and computing entropy over the resulting cluster distribution. In practice, this requires pairwise semantic comparisons between generated answers, commonly implemented using a bidirectional entailment model to determine whether two outputs belong to the same semantic group.

Our approach differs in that it does not perform semantic clustering or entailment-based equivalence checks. Instead, we embed all sampled outputs into a continuous embedding space and quantify uncertainty directly from the geometric dispersion of these embeddings. As a result, uncertainty is computed without discrete cluster assignments or auxiliary semantic matching models.

\subsection{The Directional Concentration Measure}

Given an input $x$, and an LLM $f$ $N$ sampled model outputs were obtained by passing $x$ through $f$ $N$ times. Let those output be denoted by
\[
\mathcal{O}(x) = \{{y_1, y_2, \ldots, y_N}\}
\]
Here each $y_i$, $i=1,2,\ldots, N$ represent a sample from the output distribution of $f$

Following generation of $y_i$ they were then mapped to a $d$-dimensional vector using a pretrained  embedding model $\phi(\cdot)$. The resulting embeddings are $\ell_2$-normalized,
\[
\mathbf{z}_i = \frac{\phi(y_i)}{\|\phi(y_i)\|}, \quad \mathbf{z}_i \in \mathbb{S}^{d-1},
\]
to ensure they all lie on a unit hypersphere.

Once the normalized embeddings were obtained we modeled them using the vMF distribution~\cite{mardia20}. The vMF distribution is defined on the unit hypersphere $\mathbb{S}^{d-1}$ and is parameterized by a mean direction $\boldsymbol{\mu} \in \mathbb{S}^{d-1}$ and a concentration parameter $\kappa \geq 0$. Its probability density function is given by
\[
p(\mathbf{z} \mid \boldsymbol{\mu}, \kappa)
= C_d(\kappa)\exp(\kappa \boldsymbol{\mu}^\top \mathbf{z}),
\]
where
\[
C_d(\kappa) = \frac{\kappa^{d/2 - 1}}{(2\pi)^{d/2} I_{d/2 - 1}(\kappa)}
\]
and $I_\nu(\cdot)$ denotes the modified Bessel function of the first kind.

The parameter $\kappa$ quantifies the degree of concentration of the distribution around the mean direction $\boldsymbol{\mu}$, with larger values indicating more tightly clustered samples and smaller values indicating greater dispersion. In our setting, the dispersion of the fitted distribution reflects the variability across $\mathcal{O}(x)$, and was therefore used to characterize response uncertainty.

The parameters of the vMF distribution are estimated via maximum likelihood estimation (MLE).
 Let
\[
\mathbf{R} = \sum_{i=1}^N \mathbf{z}_i
\]
denote the resultant vector, and let $\bar{R} = \|\mathbf{R}\| / N$ denote the mean resultant length. The maximum likelihood estimate of the mean direction is given by
\[
\hat{\boldsymbol{\mu}} = \frac{\mathbf{R}}{\|\mathbf{R}\|}.
\]
The concentration parameter $\kappa$ is obtained by solving
\[
A_d(\kappa) = \bar{R},
\]
where $A_d(\kappa) = \frac{I_{d/2}(\kappa)}{I_{d/2 - 1}(\kappa)}$. This equation was solved numerically using Newton-Raphson iterations, with a bisection-based fallback to ensure stability.

The concentration parameter $\kappa$ measures how tightly the embeddings cluster around the estimated mean direction. We use the inverse concentration, $\kappa^{-1}$, as the uncertainty score, with lower concentration corresponding to higher uncertainty.

Figure~\ref{fig:dcu_ex} demonstrates the DCU metric for two independent LLM queries from the dataset TriviaQa. LLaMa-2 was used for generations and the embeddings were obtained using the e5-large-v2 model. The radial plots indicate the spread of normalized  embeddings around the estimated mean direction. Note that the spread of embeddings are much higher for query 2 compared to query 1 which which reflects lower directional concentration and consequently higher uncertainty in the model’s responses.
\section{Experiments and Results}

This section describes the experimental protocol used to evaluate our DCU measure and to compare it against SE. Our general goal is to determine the viability of DCU as it relates to SE, the most similar sampling-based uncertainty method for generated text, while demonstrating our method's flexibility and improved generalization. We compare DCU to SE directly on models and datasets previously reported in \citet{farquharetal24}, and then new experiments on visual question answering datasets which showcase the performance differences between methods on a more challenging task.

Because our goal is to study uncertainty under free form generations, our experiments were designed across multiple question-answering (QA) datasets and open source LLMs for efficient benchmarking. Due to resource constraints, smaller-parameter models were used in lieu of state-of-the-art models. In the coming sections, we provide detailed descriptions of (i) the datasets and sampling procedure used to construct evaluation sets, (ii) the generation and prompting configuration used to obtain multiple outputs per input, (iii) the embedding models used to represent outputs in a continuous space, (iv) the uncertainty metrics computed (DCU and SE), and (v) the evaluation methods used to assess uncertainty quality.

\subsection{Datasets and Sampling Strategy}

To ensure a robust evaluation of our proposed metric we performed evaluation on a variety of QA datasets with varying levels of complexity. Our chosen datasets consisted of both unimodal (text only) and multimodal (text+image) entries allowing us to evaluate our methods on different variety of QA tasks and also compare our approach with existing methods of computing uncertainty for free-form natural language tasks.

For text-only QA tasks, we considered three datasets: SQuAD~\cite{squad}, TriviaQA~\cite{triviaqa}, and NQ-Open~\cite{nqopen}. Each of these datasets were used by prior work \citep{farquharetal24} when evaluating the effectiveness of SE for open-ended QA tasks. SQuAD consists of questions paired with a context passage, where answers correspond to short text spans from the  given passage. TriviaQA consists of fact-based questions that require external world knowledge and are paired with one or more annotated answer strings. NQ-Open is built from real user online search queries and consists of open ended questions along with short answer annotations. All three datasets are commonly used benchmarks for evaluating QA tasks. 

Next, to evaluate our approach in a multimodal setting, we used ScienceQA~\cite{scienceqa}. ScienceQA is a multiple choice scientific question answering dataset consisting of simple scientific questions across a variety of domains, which gives us the opportunity to evaluate on (i) a more challenging task, given the multimodal aspect, and (ii) a dataset that has not previously been reported in work on uncertainty, in particular on work using semantic entropy. One issue with ScienceQA, as it pertains to our experiments, is that it is not free-form QA. Indeed, there are few open-ended visual QA datasets commonly used in benchmarking. In our implementation details, we explain how we alter ScienceQA to allow experimental comparisons for multimodal models. Each question is paired with a set of candidate answer options and may additionally include auxiliary information such as hints, lecture style explanations, solutions, and/or an associated image. The dataset is designed to support both text only and multimodal reasoning, with questions covering topics such as biology, physics and chemistry.

For our experiments we sampled a total of 300 questions from each of TriviaQA, NQ-Open and SQuAD using a sampling protocol similar to \citet{kuhnetal23}, to ensure a fair comparison between our approach and SE. For ScienceQA, we sampled 150 questions that include an accompanying image to enable experiments on complex QA tasks.

\subsection{Implementation Details}

We now discuss our experimental strategy for evaluating the DCU metric on free-form language generation tasks using our chosen model and datasets. Our setup ensures a comprehensive evaluation of our proposed metric and also an efficient comparison with the SE approach described in \cite{kuhnetal23}. 

\subsubsection{Models}

For our experiments we used three open source LLMs: LLAMA-2-7B, Falcon-7B and Mistral-7B, which were three of the models used by \citet{farquharetal24} in prior experiments. Similarly, we experiment on three datasets --- TriviaQA, SquAD and NQ-Open --- each of which is reported on in prior experiments. For our multimodal experiments on ScienceQA, we used LLaVA 1.5 7B and Gemma 3 4B. The models were used in their pretrained forms with no additional fine-tuning.

For the implementation of DCU, an embedding model is required. For our experiments, which embed text from unimodal models, we used e5-large-v2 as our embedding model with experiments involving Llama, Falcon, and Mistral. For ScienceQA, given the multi-modal nature of the dataset and task, we used the bi-modal CLIP encoder to generate text embeddings. Similarly to the generative models the encoder models were used in their pretrained forms with no additional fine tuning. For SE, we repeat the authors' prior implementation, using a pretrained MNLI model (Deberta-Large-MNLI)\footnote{\url{https://huggingface.co/microsoft/deberta-large-mnli}} to compute pairwise semantic equivalences using bidirectional entailment.

\subsubsection{Prompting and Generation}

Our prompting strategies were designed to ensure that all models generate free-form natural language answers with no restrictions. For TriviaQA and NQ-Open, each prompt consisted of the question text and instructs the model to produce a complete textual answer. For SQuAD, each prompt included both the context passage and the associated question for a proper free form answer.

For ScienceQA, each prompt included the question text together with the set of multiple choice answer options. However, instead of asking the model to directly select an option, we reformulated the task into a free form generative QA setting by forcing the model to generate a free form answer through structured prompting. Our prompt enforced a structured format requiring the model to (i) state the chosen option using its full textual form, (ii) provide reasoning, and (iii) contrast the selected option with incorrect alternatives. This allowed us to convert the original multiple choice questions into free form answers suitable for our experiments.

It is important to note that many questions ScienceQA consists of a form of context labelled as `solution' which essentially lists the correct option and provides an explanation. For all such questions `solution' was excluded from the prompt to avoid leakage of the worked answer into the model input.


For our experiments, we used $N=10$ generations with fixed `temperature' and `top\_P' settings for each model. This is a similar setting to prior works, and has been found to be a stable balance between model performance and uncertainty calibration. Thus, we reuse it for our experiments.
For a given question the prompt remained unchanged across the $N$ generations, and the hyperparameter settings were kept fixed across datasets for efficient comparison. 

\subsubsection{Performance Evaluation and Baseline}

Following \cite{kuhnetal23}, we evaluated our uncertainty metric using both accuracy and  the area under the receiver operating characteristic curve (AUROC). AUROC measures how well the uncertainty score separates correct and incorrect model outputs across different confidence thresholds. AUROC lies between 0 and 1 with scores close to 1 indicating better predictions, and 0.5 indicating the baseline of a random prediction.

For TriviaQA, SQuAD, and NQ-Open, as a part of the dataset every question has an associated reference answer. The correctness labels for computing AUROC were determined by comparing the model’s first generated answer to the given reference answer using a ROUGE-L F1 similarity score. If the score exceeded a threshold of 0.1, the answer was treated as correct; otherwise, it was treated as incorrect.


For ScienceQA, the reference answer were in the form of an option, while our prompting strategy produced free form explanatory responses. As a result ROUGE-L F1 score proved unsuitable for determining the correctness labels for computing AUROC. To address this, we determined the correctness levels for AUROC using a CLIP encoder based semantic matching procedure. For each question we took the model's first generated answer, embedded it with the CLIP text encoder. After this we embedded all  multiple choice answer options in the same space. Using the embeddings we computed a cosine similarity between the embedding of the first generated answer and each multiple choice embedding, The option with the highest similarity was selected as the model's inferred choice. Finally if the inferred choice index matched the ground-truth option index, the response was labeled as correct; otherwise, it was labeled as incorrect. 

To estimate variability across all  experiments,  we applied a non-parametric bootstrap resampling during evaluation process rather than during model inference. After model outputs, correctness labels, and uncertainty scores were computed, the questions were repeatedly resampled with replacement and accuracy and AUROC were recomputed on each resampled set. This allowed us to obtain variability estimates for both accuracy and AUROC without requiring any additional inference or uncertainty computation.

\section{Results and Discussion}

We now present the empirical results comparing our proposed uncertainty metric against SE. We begin with the datasets that allow direct comparison to SE in terms of the authors' original experiments, involving relatively simple QA tasks. Next, we show the results of experiments on a visual question answering task.

\subsection{Application to Simple QA Tasks}

We begin by evaluating our proposed Directional metric on the three simple question answering datasets: NQ-Open, SQuAD, and TriviaQA. 

Table~\ref{tab:results_simpqa} presents the accuracy and AUROC scores across the three text only QA benchmarks (NQ-Open, SQuAD, and TriviaQA). Overall, both DCU and SE yield meaningful results, with SE generally achieving slightly higher AUROC values across most models and datasets, with differences falling within the confidence intervals derived through bootstrapping. However, DCU remains competitive, and in several settings (e.g., Falcon on NQ-Open and SQuAD), DCU attains AUROC scores that closely match or exceed the corresponding SE results.

The accuracy varies significantly across the three datasets, with NQ-Open being the least accurate compared to SQuAD and TriviaQA. In contrast, AUROC values remain relatively stable across datasets, suggesting that DCU can recover informative uncertainty signals even when absolute accuracy is low. Finally, bootstrap resampling shows that both DCU and SE exhibit low variance and tight 95\% confidence intervals across all benchmarks and models, indicating that the observed differences in AUROC are robust and not driven by sampling noise.

Our experiments indicate that DCU with embeddings is able to capture the same signal that enabled the success of SE. In other words, entropy of semantically defined clusters of texts is similar to our representational concentration measure, particularly in terms of capturing model uncertainty.

\begin{table*}[!ht]
\centering
\caption{Bootstrap aggregated results ($N{=}1000$) for TriviaQA, SQuAD and NQ-Open including accuracy, AUROC(DCU), and AUROC(SE).}
\label{tab:results_simpqa}
\begin{tabular}{llccc}
\toprule
Dataset & Model & Accuracy & AUROC$_{DCU}$ & AUROC$_{SE}$ \\
\midrule
 & LLaMA-2 & $0.46 \pm 0.06$ & $0.61 \pm 0.07$ & $0.71 \pm 0.06$ \\
TriviaQA & Mistral & $0.40 \pm 0.06$ & $0.56 \pm 0.07$ & $0.67 \pm 0.06$ \\
 & Falcon & $0.56 \pm 0.06$ & $0.68 \pm 0.06$ & $0.72 \pm 0.06$ \\
\midrule
 & LLaMA-2 & $0.91 \pm 0.03$ & $0.67 \pm 0.11$ & $0.74 \pm 0.10$ \\
SQuAD & Mistral & $0.87 \pm 0.04$ & $0.75 \pm 0.07$ & $0.82 \pm 0.07$ \\
 & Falcon & $0.82 \pm 0.05$ & $0.72 \pm 0.07$ & $0.66 \pm 0.08$ \\
\midrule
 & LLaMA-2 & $0.18 \pm 0.04$ & $0.61 \pm 0.09$ & $0.62 \pm 0.08$ \\
NQ-Open & Mistral & $0.16 \pm 0.04$ & $0.63 \pm 0.08$ & $0.58 \pm 0.08$ \\
 & Falcon & $0.31 \pm 0.05$ & $0.75 \pm 0.06$ & $0.67 \pm 0.07$ \\
\bottomrule
\end{tabular}
\end{table*}

\subsection{Application to visual QA tasks} \label{sec:results:visual}

The goal of previous experiments were to confirm the viability of DCU in relation to the prior state-of-the-art. The core of our argument is that a representational approach that matches prior baselines has the benefit of generalization to any task or output modality, without the reliance on specialized notions of textual equivalence. We found this to be the case, and in fact additionally found that our approach may prove superior to SE in some cases. Next, we present results of our experiments on the performance differences of DCU and SE on a more challenging task, visual QA.


\begin{table*}[!htbp]
\centering
\caption{Bootstrap aggregated results ($N{=}1000$) for ScienceQA including Accuracy, AUROC(DCU), and AUROC(SE).}
\label{tab:sciqa_all}
\begin{tabular}{llccc}
\toprule
Dataset & Model & Accuracy & AUROC$_{DCU}$ & AUROC$_{SE}$ \\
\midrule
ScienceQA & Gemma-3-4B-IT & $0.67 \pm 0.07$ & $0.56 \pm 0.09$ & $0.52 \pm 0.1$ \\
ScienceQA & LLaVA-1.5-7B & $0.59 \pm 0.07$ & $0.67 \pm 0.08$ & $0.51 \pm 0.08$ \\
\bottomrule
\end{tabular}
\end{table*}

Table~\ref{tab:sciqa_all} shows results of our experiments on ScienceQA on two multimodal models. A number of findings emerge from these results. First, the overall AUROC scores for SE and DCU are low, especially for Gemma 3, both in relation to text-only datasets (range of AUROC was 0.6-0.8) and in general, as an AUROC score of 0.5 indicates the predictions are no better than random. Second, the AUROC scores for DCU with embeddings are clearly superior to SE especially for LLaVA. In general, the scores ($0.67$ for LlaVA) are good, but not acceptable AUROC levels, showing that there is still room for improvement.

These findings are both an indication of the limits of SE, with poor performance on data and tasks outside the typical QA datasets used in unimodal model benchmarking (and in prior evaluations of the method), and the advantages of the representational DCU method.
\section{Conclusion}

In this work, we have argued and empirically demonstrated the limits of semantic clustering methods, a key baseline in UQ for LLMs. These limits are highlighted by the complexity of textual clustering, and in particular the rigidity of the notion of ``semantic'' uncertainty and clustering sampled texts based on their semantic equivalence. We proposed DCU, which leverages the core concept of semantic uncertainty and other, similar sampling-based methods, which is to quantify uncertainty as the inconsistency of sampled responses. In our experiments, we showed its strong performance in comparison to baseline question-answering, as well as more complex tasks such as visual question answering.

The implication of our work is that the underlying reason for the success of previous sampling-based methods, particularly semantic entropy, can be accessed through embeddings and our proposed DCU metric. By relying only on embeddings, we remove the need for semantic equivalence from the framework, which had limited the applicability to textual domains that can be meaningfully divided by textual semantics. This clearly breaks down outside the realm of simple question-answering, as most tasks and generative model outputs are not so simply categorized.

\subsection{Future Work}

The only requirement for DCU is a good embedding model for the domain and modality at hand. As a result, it can be applied to any generative model and any data domain (e.g., image generation or multi-modal outputs), provided that there exists a good enough encoder. This is the first key area we suggest for future work, exploring the ways that uncertainty can be quantified in arbitrary modalities (e.g., image or audio generation) as well as different textual tasks beyond question answering.

The second area for future work that we identify is the integration of our core methodology into agentic systems. In particular, recent work has investigated the phenomenon of propagating uncertainty in this setting \citep{duan2025uprop}. Because of the flexibility of embedding-based methods, which require no prior assumptions about modality or task, DCU can be injected into agentic pipelines, acting on hidden states of samples from a model to detect and limit uncertainty before propagating to further steps.

\subsection{Limitations}

While we have achieved great flexibility in our framework for UQ for generative models, we are still limited by some of the same factors as prior work. Namely, our method still requires repeated samples from a model. This is a costly requirement that may restrict application of our method in real-world settings. However, the computational cost of encoder models is lower than that of generative models, which were required by prior methods. Further empirical investigation must determine whether this is still a barrier for real-world applications, or whether further improvements are necessary.

Lastly, further empirical investigation is planned which will confirm the viability of the DCU framework across myriad modalities and tasks. Our initial experiments indicate its viability in such settings, and future work will confirm and document these capabilities.

\section{Acknowledgments}

This work was supported by the NNSA Office of Defense Nuclear Nonproliferation Research and Development, U.S. Department of Energy, and Pacific Northwest National Laboratory, which is operated by Battelle Memorial Institute for the U.S. Department of Energy under Contract DEAC05–76RLO1830. This article has been cleared by PNNL for public release as PNNL-SA-219787.

\bibliographystyle{icml2026}
\bibliography{bibliography}

\clearpage
\appendix

\section{Appendix}
\subsection{Derivation of the Directional Concentration Parameter}
Let $x$ be a fixed input/query to a LLM and $\mathcal{O}(x) = \{y_1,\dots,y_N\}$ denote the $N$ stochastic outputs. 
Using a pretrained encoder each output $y_i$ is embedded into $\mathbb{R}^d$ via $\mathbf{v}_i = \phi(y_i)$ and normalized as:
\begin{equation}
    \mathbf{z}_i = \frac{\mathbf{v}_i}{\|\mathbf{v}_i\|}, \qquad \mathbf{z}_i \in \mathbb{S}^{d-1}. \label{eq:z_i_def}
\end{equation}
to lie on a unit hypersphere with intrinsic $d-1$ dimensions.

$\mathbf{z}_1,\dots,\mathbf{z}_N$ modelled as i.i.d.\ draws from a $\mathrm{vMF}(\boldsymbol{\mu},\kappa)$ distribution with mean direction $\boldsymbol{\mu}\in\mathbb{S}^{d-1}$ and concentration $\kappa\ge0$. The density function for $\mathbf{z}_1,\dots,\mathbf{z}_N$ is then defined as:
\begin{equation}
    p(\mathbf{z}\mid\boldsymbol{\mu},\kappa)
    = C_d(\kappa)\exp\!\left(\kappa\,\boldsymbol{\mu}^\top \mathbf{z}\right), \label{eq:vmf_density}
\end{equation}
where
\begin{equation}
    C_d(\kappa)
    = \frac{\kappa^{d/2 - 1}}{(2\pi)^{d/2} I_{d/2 - 1}(\kappa)}, \label{eq:normalizing_constant}
\end{equation}
and $I_\nu(\cdot)$ denotes the modified Bessel function of the first kind.

\paragraph{Log Likelihood:}
Given the normalized embeddings $\{\mathbf{z}_i\}_{i=1}^N$, the log likelihood function parameterized on $(\boldsymbol{\mu},\kappa)$ can be defined is
\begin{equation}
    \ell(\boldsymbol{\mu},\kappa)
    = \sum_{i=1}^N \left[\log C_d(\kappa) + \kappa\,\boldsymbol{\mu}^\top \mathbf{z}_i \right]. \label{eq:loglik_pre}
\end{equation}
 The resultant vector $R$ is now defined as:
\begin{equation}
    \mathbf{R} = \sum_{i=1}^N \mathbf{z}_i. \label{eq:R_def}
\end{equation}
and let $\|\mathbf{R}\|$ denote the norm of R. 

Thus Substituting \eqref{eq:R_def} into \eqref{eq:loglik_pre} yields
\begin{equation}
    \ell(\boldsymbol{\mu},\kappa)
    = N\log C_d(\kappa) + \kappa\,\boldsymbol{\mu}^\top \mathbf{R}, \label{eq:loglik_mu_kappa}
\end{equation}

The goal here is to maximize $\ell(\boldsymbol{\mu},\kappa)$ with respect to $\mu$ and $\kappa$ with the constraint $\|\boldsymbol{\mu}\| = 1.$. Thus the final log likelihood becomes
\begin{equation}
    \mathcal{L}(\boldsymbol{\mu}, \kappa)
    = \ell(\boldsymbol{\mu},\kappa) + \lambda\left(1 - \boldsymbol{\mu}^\top\boldsymbol{\mu}\right). \label{eq:lagrangian}
\end{equation}
where $\lambda$ is the Lagrange multiplier.

\paragraph{Estimation of $\mu$}

The score equation for estimating $\mu$ can be then obtained as
\begin{equation}
    \frac{\partial \mathcal{L}}{\partial\boldsymbol{\mu}}
    = \kappa \mathbf{R} - 2\lambda \boldsymbol{\mu} = \mathbf{0}. \label{eq:lagrange_eq}
\end{equation}
Solving \eqref{eq:lagrange_eq} for $\boldsymbol{\mu}$ yields
\begin{equation}
    \boldsymbol{\mu} = \frac{\kappa}{2\lambda}\mathbf{R}. \label{eq:mu_parallel}
\end{equation}

Since $\|\boldsymbol{\mu}\| = 1$, substituting \eqref{eq:mu_parallel} in place of $\mu$ yields
\begin{equation}
    \hat{\boldsymbol{\mu}} = \frac{\mathbf{R}}{\|\mathbf{R}\|}, \qquad \text{if } \|\mathbf{R}\| \neq 0. \label{eq:mu_hat}
\end{equation}

\paragraph{Estimation of $\kappa$}
Upon substituting $\hat{\mu}$ into \eqref{eq:loglik_mu_kappa} the log likelihood becomes
\begin{equation}
    \ell(\kappa) = N\log C_d(\kappa) + \kappa \|\mathbf{R}\|. \label{eq:loglik_profile}
\end{equation}

Then the score equation for $\kappa$ becomes:
\begin{equation}
    \frac{d\ell}{d\kappa}
    = N\frac{d}{d\kappa}\log C_d(\kappa) + \|\mathbf{R}\|. \label{eq:dl_dkappa}
\end{equation}
By substituting $\log C_d(\kappa)$ from \eqref{eq:normalizing_constant} the equation becomes,
\begin{equation}
    \log C_d(\kappa)
    = \left(\tfrac{d}{2}-1\right)\log\kappa
      - \tfrac{d}{2}\log(2\pi)
      - \log I_{d/2 - 1}(\kappa). \label{eq:logC}
\end{equation}

Now, using  Bessel derivative identities and recurrence relations:
\begin{equation}
    \frac{d}{d\kappa}\log I_{\nu}(\kappa)
    = \frac{I_{\nu+1}(\kappa)}{I_{\nu}(\kappa)} + \frac{\nu}{\kappa}, \label{eq:bessel_identity}
\end{equation}
and substituting $\nu = d/2 - 1$ into \eqref{eq:bessel_identity}, the score equation further simplifies to
\begin{equation}
    \frac{d}{d\kappa}\log C_d(\kappa)
    = -\frac{I_{d/2}(\kappa)}{I_{d/2 - 1}(\kappa)} = - A_d(\kappa). \label{eq:dlogC}
\end{equation}

Substituting $A_d(\kappa)$ into \eqref{eq:dl_dkappa} gives the score equation reduces to
\begin{equation}
    \frac{d\ell}{d\kappa}
    = -N A_d(\kappa) + \|\mathbf{R}\|. \label{eq:score}
\end{equation}
Setting \eqref{eq:score} to zero yields
\begin{equation}
    A_d(\hat{\kappa}) = \frac{\|\mathbf{R}\|}{N} = \bar{R}. \label{eq:likelihood_equation}
\end{equation}

The estimate if $\kappa$ is obtained by solving \eqref{eq:likelihood_equation} numerically using Newton iteration and bisection method is used in case of a fallback. Newton iterations may diverge or become numerically unstable when $\bar{R}$ is close to 0 or 1. In cases like this bisection always yields the unique root and thus serves as a reliable fallback when Newton's method fails to converge.

\end{document}